# Hybrid stacked ensemble combined with genetic algorithms for Prediction of Diabetes

Jafar Abdollahi [a], Babak Nouri-Moghaddam [a,b,*]

[a] Department of Computer Engineering, Ardabil Branch, Islamic Azad University, Ardabil, Iran
[b] Department of Industrial Engineering, Iran University of Science and Technology, Tehran, 1684613114, Iran

**Abstract**

Diabetes is currently one of the most common, dangerous, and costly diseases in the world that is caused by an increase in blood sugar or a decrease in insulin in the body. Diabetes can have detrimental effects on people's health if diagnosed late. Today, diabetes has become one of the challenges for health and government officials. Prevention is a priority, and taking care of people's health without compromising their comfort is an essential need. In this study, the Ensemble training methodology based on genetic algorithms are used to accurately diagnose and predict the outcomes of diabetes mellitus. In this study, we use the experimental data, real data on Indian diabetics on the University of California website. Current developments in ICT, such as the Internet of Things, machine learning, and data mining, allow us to provide health strategies with more intelligent capabilities to accurately predict the outcomes of the disease in daily life and the hospital and prevent the progression of this disease and it's many complications. The results show the high performance of the proposed method in diagnosing the disease, which has reached 98.8%, and 99% accuracy in this study.

## Introduction

Diabetes is a chronic endocrine disorder that affects the body's metabolism and causes structural changes. Since 2014, the spread of the disease has increased from 100 million to 422 million patients [1-2-3]. Diabetes is typically divided into type 1, type 2[4], and gestational diabetes, which Type 2 is increasing with a high prevalence worldwide and is one of the leading causes of death. Because regardless of age and gender, it threatens them due to the lack of insulin in the body [5]. Increased blood sugar is associated with the risk of death in the community due to pneumonia, stroke, acute myocardial infarction, etc. However, its effect is on the vital organs is so harmful that it is considered the mother of all diseases. According to the World Health Organization (WHO), most women with diabetes have no information about it. Especially in pregnant women, the disease can be transmitted to their children. There is a risk of miscarriage, kidney failure, heart attack, blindness, and other chronic and deadly diseases in diabetic women. Therefore, it is important to diagnose diabetes faster in pregnant women [6-7-8].

The promising emerging potential of the Internet of Things (IoT) for connected medical devices and sensors plays an important role in the next-generation healthcare industry for quality patient care. Due to the increasing number of elderly and disabled people, there is an urgent need for real-time health care infrastructure to analyze patient health care data to prevent preventable deaths [9]. Also, in the field of smart health, modern wearable devices have gradually increased their capabilities in recent decades and are equipped with several internal and external sensors that can detect many vital signs [10]. Designing and implementing a remote monitors system allow physicians and caregivers to be aware of peopling health at all times, and current developments in ICT such as the Internet of Things, Machine learning, and Data Mining allows us to provide health strategies with more intelligent capabilities to accurately predict the outcomes of the disease in daily life, and the hospital.

Besides, medical advances in recent decades have significantly increased life expectancy while significantly reducing mortality [11]. According to Reference [12], the benefits of personal health care with IoT requirements are divided into three general categories, such as.

Increase the likelihood of early detection of potential and ongoing diseases without the need to visit clinics. Promote frequent assessments of health conditions and awareness of preventive health care needs.

The diagnosis of the disease with a set of measurements for a period that can be been effectively diagnosed through routine examinations in clinics. In clinics, physicians first understand the symptom, then they perform medical tests and diagnose diseases using current measurement data, and a few previous measurements available.

In this study[13], IoT technology provides a correct and structured approach to improving human health, which is expected to change the IoT based devices of the healthcare sector regarding social benefits and influence, as well as cost-effectiveness, and due to the nature of IoT calculations, all institutions of the health system (people, equipment, medicine) can be continuously monitored and managed, and the use of these technologies in the healthcare industry can improve the quality and costs of medical care by automating tasks already performed by humans. Figure 2, shows the general IoT-based health monitoring system [14-15].

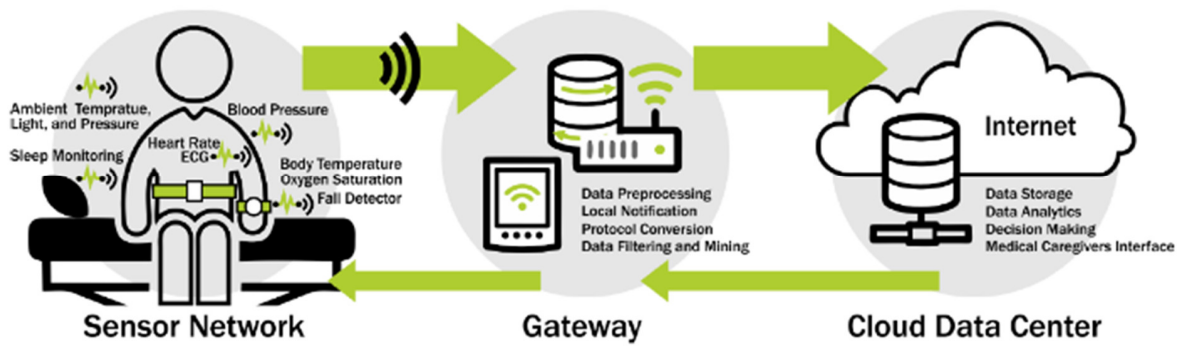

**Fig. 1.** General IoT-based health monitoring system [25]

As a result, the prediction accuracy of a model may be high even with optimal parameters. Not surprisingly, the models produced by MLR and carrier support regression with a linear core are not statistically distinct and perform significantly better than other methods in the IWPC Ensemble [55]. One way to overcome the limitations of a single algorithm is to combine the advantages of several algorithms to exceed the limit of a single machine learning algorithm (e.g., the Ensemble method). Recently, the "bagging" Ensemble method has been used to predict diabetes [58, 59].

Stack generalization is another Ensemble method that uses a higher-level model to combine lower-level models to achieve higher prediction accuracy [60, 61]. Unlike bagging and boosting approaches, which can only combine machine learning algorithms of the same type, stacked generalization can combine different types of algorithms through a meta-machine learning model to maximize generalization accuracy. The purposes of this article are as follows:

- Combining the best voting machine learning architectures
- Using a hybrid machine learning model to diagnose diabetes
- Significant improvement in a forecast accuracy
- Use several models in combination
- Achieve a high level of reliability in classification
- Use multiple models to increase the estimate of the final model
- Increased accuracy and reduced error compared to single-core models

**So the main contribution of this article:**

- Machine learning ensemble models for Diabetes (T2D) prediction demonstrated high performance.
- Comparing results with the most related researches according to the literature.

- Examining the benefits of ensemble methods proposed recently for prediction.
- Using hybrid Stacked Generalization based Metaheuristics approach in the diagnosis of Diabetes.

**The rest of the paper is as follows**: we will review the method in section two; we will examine the results in section three, we will review the discussion in the fourth section, and finally, in the last section, we will review the conclusions and future work.

## Related Work

The present studying deals with an ensemble stacking-based learning methodology for the detection of diabetes. This section provides a brief review of literature on numerous meta-heuristic optimization-based and ensemble learning-based prediction methods of diabetes. Measures have been taken to reduce the number of chronic diseases diagnostic tests to reduce overall costs. One of the possible solutions is to use machine learning techniques in healthcare data, which are used to find frequent patterns in a large database to obtain useful information.

Machine learning methods are very useful in diagnosing diabetes and increase its efficiency. The most important challenge for researchers in machine learning, and has been diagnosing diabetes a decade ago. In reference [16], researchers have studied different methods of machine learning to diagnose various diseases such as heart disease, diabetes, liver, and hepatitis and have achieved good results in diagnosing this disease. In reference [17], artificial neural networks and Bayesian networks have been used to classify diabetes and cardiovascular disease.

The purpose of this reference is to study artificial neural networks and Bayesian networks and their application in the classification of Type 2 diabetes (T2D) and cardiovascular disease. In reference [18], a very hybrid method (artificial neural network, and genetic algorithm) has been used to effectively diagnose heart disease. The proposed method can increase the performance of the genetic algorithm by almost 10% more than the performance of the neural network. In reference [19], the artificial neural network approach has been used to diagnose heart disease and specific sensitivity and specificity have been achieved with 95.95% and 95.91% accuracy, respectively.

In the reference [20], six different learning algorithms have been used to classify heart disease in MATLAB, and VEKA, including linear SVM, two-way SVM, Coby SVM, Medium SVM, Medium Gaussian SVM, and decision tree. Also, in reference to [21], a machine learning method called (PSSVM) has been used to help diagnose heart disease. In this study, BFGS methods have been used to determine the status of heart disease in patients.

ML algorithms were designed and are used for collecting medical data. Today, ML offered various tools for efficient data analysis. Especially in the last few years, the digital revolution has provided affordable and accessible tools for collecting and saving data. Data collection and examination machines are located in new and modern hospitals to be able to collect and share data in large information systems. ML technology is very effective for analyzing medical data and has an effective role in solving diagnostic problems. Correct diagnostic data is presented as a medical record or report in modern hospitals or their specific information department. These techniques are used to classify the data set [16,22].

Significant steps have been taken to estimate the appropriate dose of warfarin to improve the care of diabetic patients. Many Pharma-co-genetic algorithms have been developed to predict the required does in individual patients [54-57]. Most dosing algorithms are based on multivariate linear regression (MLR). One of the most widely used and tested algorithms is the IWPC Pharma-co-genetic Algorithm [55], which has been proven in several studies.

Other advanced machine learning methods such as deep learning (neural network), decision trees, and support vector machines have also been used to predict warfarin doses [54, 56, 57], but in these studies, a machine learning algorithm to maximize Accuracy is used. Accurate prediction of warfarin dose. Mathematically, a machine learning algorithm is a complex feature for a nonlinear function, and a machine learning model may be appropriate for a particular subset of patients but may be appropriate for other patients with genetic and racial background patients.

## Supported methodologies

This section discusses all the supported methodologies, used in this study. Machine learning techniques have been widely used in many scientific fields, but this use in the medical literature is limited partly because of technical difficulties.

- **Decision Tree**: A decision tree is a decision support tool that uses trees to model [23-24].
- **Naive Bayes classifier**: A group of simple classifiers based on probabilities are said to be based on Bayes' theorem, assuming the independence of random variables. [25, 26].
- **Artificial neural network**: Inspired by the way the biological nervous system works to process data and information for learning and knowledge creation [27-28].
- **Support vector machine**: It is one of the supervised learning methods used for classification and regression [28-29].
- **C 4.5**: Algorithm C 4.5 is one of the decision tree algorithms which is very important due to its very high interoperability [30].
- **Random forest**: One combination learning method for categorization is regression, which works Based on training time and the output of classes (classification) or for the predictions of each tree individually, based on a structure consisting of many decision trees [31,32].
- **k-nearest neighbors**: KNN classifier is to classify unlabeled observations by assigning them to the class of the most similar labeled examples. Characteristics of observations are collected for both training and testing datasets. The appropriate choice of k has a significant impact on the diagnostic performance of the KNN algorithm. A large k reduces the impact of variance caused by random error but runs the risk of ignoring small but important patterns [31,35,36].

The following figure shows the proposed flowchart [36]. In the following, we will examine the popular methods of combining categories.

- **Bagging**: One of the simplest and most successful combined approaches to improve the classification problem is the Bagging algorithm, which is commonly used for decision trees. This algorithm is very useful for bulk data and will work well for unstable learning algorithms, that is, algorithms that change as a result of changing data.
- **Boosting**: This algorithm aims to combine several weak classifiers and obtain a strong one to improve performance, in which the predictors are trained continuously.
- **Ada boost**: a meta-algorithm is designed to improve the performance and solve the problem of unbalanced categories, which produces a powerful and high-quality learner from a combination of three-week learners. To solve difficult nonlinear problems, this algorithm combines weak learners to produce an accurate classifier [37-38-39-40-41].

This study is to provide an intelligent monitoring system for patients and elderly people with chronic diseases using the collected data for effective diagnosis and prediction in non-critical situations to promote intelligent health and prevent deaths on IoT infrastructure using the intelligent ensemble learning algorithms. Ensemble methods are learning algorithms that construct a set of classifiers and then classify new data points by taking a (weighted) vote of their predictions. The original ensemble method is Bayesian averaging, but more recent algorithms include error-correcting output coding, Bagging, and boosting. This paper reviews these methods and explains why ensembles can often perform better than any single classifier.

## Material and Methods

Since the use of an intelligent machine learning algorithm has not been effective and accurate in diagnosing and predicting diseases and has not been successful in many scenarios, and considering the content and challenges mentioned in the background section, there are many challenges to smart health in IoT needed to be addressed, one of which is the accurate diagnosis and prediction of disease outcomes. In this paper, we use the new machine learning approach called Ensemble Learning for the diagnosis and prediction of chronic diseases that will be described below. The purpose of this paper is to improve the accuracy and speed of diagnosis of chronic diseases in the context of the intelligent network by which we want to use ensemble learning approaches as well as a new meta-learner in stacking learning. Stack generalization is an approach that allows researchers to connect several different prediction algorithms to a combination.

**Data Set**

in this study, the data set of type 2 diabetes available in (https://archive.ics.uci.edu /ml/datasets/ diabetes has been used, which has 9 useful variables and 768 records. These variables and abbreviations are listed in Table 1. 70% of the data is for training and 30% of the data is for testing [6

**Table 1**
Description of the Pima Indian diabetes datasets.

| Description of the Pima Indian diabetes Datasets. | | | | | |
|---|---|---|---|---|---|
| **Dataset** | Sample Size | Feature size including class label | Classes | Presence of missing attribute | Presence of noisy attributes |
| **Pima Indian diabetes** | 768 | 9 | 2 | NO | NO |

In addition, another dataset has been used to teach algorithms. This data has been prepared to analyze factors related to readmission as well as other outcomes pertaining to patients with diabetes. in this study, the data set of heart patients available in (https://archive.ics.uci.edu/ml/datasets/Diabetes+130-US+hospitals+for+years+1999-2008 has been used, which has 20 useful variables and 768 records. These variables and abbreviations are listed in Table 2. 80% of the data is for training and 20% of the data is for testing.

**Table 2**
Description of the Diabetes 130-US hospitals for years 1999-2008 Data Set

| Diabetes 130-US hospitals for years 1999-2008 Data Set | | | | | |
|---|---|---|---|---|---|
| **Dataset** | Sample Size | Feature size including class label | Classes | Presence of missing attribute | Presence of noisy attributes |
| **Diabetes 130-US hospitals for years 1999-2008 Data Set** | 100.000 | 55 | Multivariate | Yes | NO |

**Data preprocessing**

Data preprocessing is necessary to prepare the diabetes type data and Pima Indians data in a manner that a deep learning model can accept. Separating the training and testing data sets ensures that the model learns only from training data and tests its performance with the testing data. The data set was divided into training and test data. The training data contain 70% of the total data set, and the test and validation data contain 15% each. At first, all the data was shuffled.

**Building and training the Stacked-Generalization Model:**

In this paper, we develop a stacking-based evolutionary ensemble learning system "Stacked Generalization based Metaheuristics" for predicting the onset of Type-2 diabetes mellitus (T2DM) within five years. Before learning, as a data preprocessing step, the missing values and outliers identified and imputed with the median values.

For base learner selection, several machine learning optimization algorithms are utilized which simultaneously maximizes the classification accuracy and minimizes the ensemble complexity. As for model combination, Bagging, Boosting, and Ada boost are employed as a meta-classifier that combines the predictions of the base learners. The selected parameters are listed in Tables 3 and 6. The comparative results demonstrate that the proposed stacking genetic method significantly outperforms several individual ML approaches and conventional ensemble approaches. Fig. 3, depicts the learning process with Stacked Generalization based on the model selection from 6 (Table 6) base learners, and 3 stacking-based combination methods.

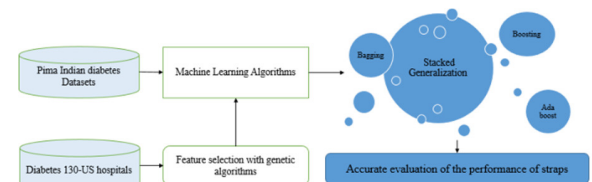

**Fig. 2.** Schematic of model proposed Framework

Stack generalization is a different technique for combining several different classifiers such as decision tree, artificial neural network, support vector machine, etc., which consists of two stages:

- basic learners at level zero and stacking model learners at level one;
- At level zero, several different models are used to learn from the dataset, and the output of each model is used to make a new dataset. Figure 3 shows the Stacking algorithm [42].

**Input**: Data set D = {(Xi, y1), (X2, y2), ..., (Xm, Ym) };
First-level learning algorithms L1, ..., $C_T$;
Second-level learning algorithm L.
Process:
 for t = 1, ... ,T:
 $h_t = L_t (D)$     % Train a first-level individual learner hy by applying the first-level
                % learning algorithm Lt to the original data set D
end;
D' = θ;     % Generate a new data set
for i = 1, ... , m:
for t = 1, ... ,T:
$z_{it} = h_t(X_i)$ % Use ht to classify the training example $X_i$
     end;
D' = D' U {(zi1, Zi2, ..., ZiT) ,yi)}
end;
h' = L(D'). % Train the second-level learner h' by applying the second-level
 ⁒learning algorithm L to the new data set D'
**Output**: H (X) = h' (h₁ (x), ... ,$h_T$ (x))

**Fig. 3.** Stacking algorithm

**Hyper parameter tuning**

Algorithms: an overview

Figure 4, shows a flow chart of the algorithms we applied in this study. The overall architecture of the diabetes (T2D) data classification model is depicted in Figure 3. Here, the Pima Indian diabetes dataset is considered to test all the models. The source of this dataset is the UCI repository. The attributes of this dataset contain the following information of a pregnant woman, such as the number of times a woman is pregnant, the concentration of glucose, thickness of skinfold, blood pressure rate, insulin rate, body mass index (BMI), and diabetes pedigree function including patient's age [43-44-45-46-47]. To determine whether ensemble predictors constructed using stacked generalization improve the prediction accuracy for warfarin stable dose, we constructed different stacked generalization frameworks using the exact same parameters in individual algorithms (Table 3, 4, 5, and 6).

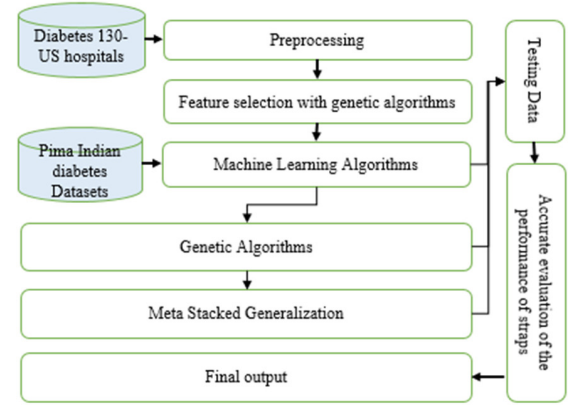

**Fig. 4.** Proposed of research

**Table 3**
Genetic algorithms parameters.

| Parameter | Value |
|---|---|
| NVAR | 9 |
| PRECI | 20 |
| NIND | 20 |
| MAXGEN | 100 |
| MIGR | 0.2 |
| INSR | 0.95 |
| SUBPOP | 5 |
| MIGGEN | 20 |

In Table 3 NVAR is the number of variables; PRECI—precision of binary representation; NIND—number of individuals; MAXGEN—maximum number of generations; MIGR—migration rate; INSR—insertion rate; SUBPOP—number of subpopulations; MIGGEN—number of generation, after which migration takes place between subpopulations. while the type of genetic operators is as listed in Table 4.

**Table 4**
Genetic operators.

| Operator | Type |
|---|---|
| Encoding | Binary |
| Reinsertion | Fitness based |
| Crossover | Double point |
| Mutation | Bit inversion |
| Selection | Roulette wheel selection |
| Fitness function | Linear ranking |

**Table 5**
hyper parameter tuning Machine Learning Algorithms.

| Technique | hyper parameter |
|---|---|
| RF | N estimators=100, criterion= *entropy* , max depth=10 , max features= n features , n classes_=2 |
| KNN | N neighbors=5, weights= uniform**,** algorithm= auto**,** metric= minkowski |
| MLP | Hidden layer sizes=(100,) , activation= relu **,** solver= adam **,** batch size=100, learning rate= ***adaptive,*** max iter =100 |
| Ada Boost | Base estimator=Decision Tree Classifier(max depth=1), n estimators=100, learning rate=1 , algorithm= *SAMME.R***,** n classes=2, |
| D Tree | Criterion= entropy**,** splitter= best**,** max depth=3, max features= auto |
| NB | Priors= n classes**,** epsilon= float**,** sigma= n classes |
| GBC | Loss= deviance**,** learning rate=0.1, n estimators=50, criterion= friedman mse, |
| SVM | Kernel= sigmoid **,** degree=3, gamma= scale |
| Extra Trees | N estimators=50, criterion=gini **,** max depth=3, max features=auto |

In Table 5 - RF is the Random Forest; KNN—k-nearest neighbors algorithm; MLP — multilayer Perceptron; Ada Boost — Ada Boost; D Tree — decision tree algorithm; NB — Naive Bayes; GBC — gradient boosting classifier algorithm; SVM — Support vector machine, Extra Trees; Extremely Randomized Trees Classifier.

## Performance measuring attributes

For the study, Jupyter notebook was used for implementation, and Python the programming language was used for coding. To evaluate the methods used and to determine how one of the available models can be considered, a model that has the most predictive accuracy compared to the other methods are compared to the categorization methods and found the appropriate one. And efficiently in this article, we have used cost-benefit analysis (the disruption matrix), ROC curve, and other model selection issues such as accuracy and so on.

performance measurement is used to determine the effectiveness of the classification algorithm so that, in the case of two-dimensional classification problems, one can show the cost of classification with a cost matrix for two types of false positive (FP) and false-negative (FN) errors and two types of classification into the positive true (TN) and negative true (TN) that give different costs and benefits. As shown in Tables 6 and 7 [6-48].

**Table 6**
Confusion Matrix

| Confusion Matrix | | Classified As: | |
|---|---|---|---|
| | | Negative | Positive |
| **Actual Class** | Negative | TN | FP |
| | Positive | FN | TP |

**Table 7**
Detail descriptions about the performance Measures [6-48].

| Performance Measure | Description |
|---|---|
| TP | When the positive samples are classified accurately |
| TN | When the negative samples are classified accurately |
| FP | When the negative samples are misclassified |
| FN | When the positive samples are misclassified |
| Accuracy | It is the overall classification accuracy percentage resulted from a standard classifier. |
| Sensitivity(Sen) | It determines the proportion of true positive samples in total samples and called as True Positive Rate (TPR) Sn=TP/(TP+FN) |
| Specificity(Spe) | It identifies the proportion of true negative samples in total samples and called as False Positive Rate (FPR) Sp=TN / (TN + FP) |
| F-score | F-score value gives the combined performance of the two classes F-score= (2 * Sp* Sn) / (Sp + Sn) |
| Roc | It is a graphical representation between sensitivity and specificity. ROC plots RPR versus FPR at different classification thresholds. |
| AUC | It is an aggregate evaluation of performance across all possible classification thresholds. So, it is called as Area under the ROC curve. |

## Results and Discussion

During the past years, medical service providers have always manually examined the vital signs of patients and diagnosed and predicted the disease based on patient records and their research findings. In this study, intelligent machine-learning algorithms are used to effectively diagnose and accurately predict the outcomes of the disease, in which cases such as age, gender, blood pressure, cholesterol, smoking, etc. are considered in the diagnosis of this disease, and finally, the risk of the disease against the mentioned diseases is determined. The table below compares the similar works of others with ours.

**Table 8**
Comparison of work results with previous works

| Authors and year of research presentation | Selective method | Model accuracy | Implementation This Paper |
|---|---|---|---|
| Singh, P. P., Prasad, S., Das, B., Poddar, U., & Choudhury, D. R. (2018).[49] | Neural network method | 88.5% | 93% |
| Bhuvaneswari, G., & Manikandan, G. (2018).[50] | Feature selection based on genetic algorithm | Improving decision accuracy 82.3 | 83% |
| Enrique V. Carrera et al.[51]. | Decision tree and support vector machine | Svm : 80.4% and 94.6% respectively and decision tree is 91.0% and 88.1% respectively | Svm : 94% Decision tree 93% |
| Emrana Kabir Hashi et al.[52] | Use machine learning algorithms | C4.5 and KNN were 90.43% and 76.96% accurate. | KNN : 94% |

The success of an Ensemble learning system is based on a variety of classifiers that make it up. If all classifiers present the same output, it is not possible to correct a possible error. So, they are more likely to have different errors on different samples. If each classifier presents a different error, you can reduce the total error after their strategic combination. So, such a set of classifiers must be diverse. This diversity can be achieved in different ways as shown in Figure 5 [53].

- Use different training datasets is to train classifiers.
- Use different training parameters is for different classifiers.
- Use different classifiers.

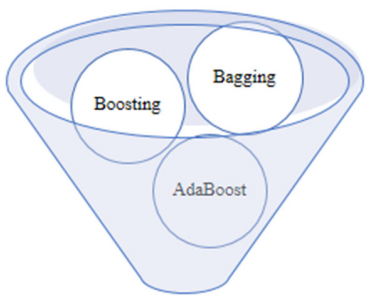

**Fig. 5**. useful methods of Ensemble learning

Today, due to the lack of knowledge about how to use different data around us, they are neglected by managers, while if these seemingly insignificant data are purposefully stored and then mined, it will generate a lot of knowledge and help us is in managerial decisions. In this study, the genetic algorithm with logistic regression is used to select the appropriate feature is based on the correct diagnosis of the desired class, after applying statistical and probabilistic approaches in the data set and also preprocessing to remove redundant and lost data to extract features that have more variance in the complications of diabetes.

For this purpose, we used a logistic regression algorithm and a random forest to calculate the accuracy and examine the features with more variance in the rapid diagnosis of diabetes. We used these two algorithms as evaluation functions to calculate the data set properties. We applied the data set properties as input to both algorithms, then this algorithm calculated the accuracy of each feature. According to the results, logistic regression has a higher performance than the random forest classifier 0.70 times with 93% accuracy and 5 optimal features.

In table 9 - Features selection for classification model attempt to select a minimally sized subset according to the following criteria: (1) The classification accuracy should increase; (2) the values for the selected features should have to close as a possible to the original class distribution. The results of feature selection on the diabetes dataset be shown in the table below.

**Table 9**
Features selection for classification model

| Parameter | DBP | TSFT | SI | BMI | DPF | AGE | PG |
|---|---|---|---|---|---|---|---|
| Value | 0.98 | 0.96 | 0.94 | 0.97 | 0.84 | 0.96 | 0.89 |

**Table 10**
Comparison of accuracy rates. Diagnosis of detection Diabetes was compared in rates among various algorithms with Holdout approach.

|  | Pima Indian diabetes datasets | | | Diabetes 130-US hospitals for years 1999-2008 Data Set | | |
| --- | --- | --- | --- | --- | --- | --- |
|  | Accuracy | Sensitivity | Specificity | Accuracy | Sensitivity | Specificity |
| RF | 93 | 95 | 93 | 95 | 94 | 92 |
| KNN | 83 | 92 | 82 | 88 | 89 | 92 |
| MLP | 90 | 85 | 88 | 93 | 90 | 89 |
| Ada boost | 92 | 94 | 88 | 86 | 88 | 87 |
| D tree Classifier | 93 | 90 | 96 | 94 | 89 | 92 |
| NB | 83 | 85 | 77 | 88 | 79 | 83 |
| GBC | 80 | 95 | 66 | 89 | 92 | 88 |
| SVM | 70 | 65 | 72 | 80 | 79 | 78 |
| Extra Tree | 80 | 86 | 79 | 91 | 93 | 92 |
| Suggest Method (ST-GA) | 98 | 97 | 100 | 99.06 | 99 | 100 |

**Table 11**
Comparison of accuracy rates. Diagnosis of detection Diabetes was compared in rates among various algorithms with K-Fold Cross-Validation approach.

| Model | Pima Indian diabetes datasets | | | Diabetes 130-US hospitals for years 1999-2008 Data Set | | |
| --- | --- | --- | --- | --- | --- | --- |
|  | K fold=5 | K fold=10 | K fold=15 | K fold=5 | K fold=10 | K fold=15 |
| RF | 87 | 93 | 90 | 93 | 94 | 95 |
| KNN | 83.26 | 94 | 91 | 88.06 | 87 | 88 |
| MLP | 89 | 90 | 89 | 92 | 93 | 93 |
| Ada boost | 50 | 52 | 78 | 87 | 86 | 86 |
| D tree Classifier | 78 | 95 | 95 | 93 | 93.56 | 94 |
| NB | 55 | 55 | 55 | 86 | 89 | 88 |
| GBC | 73 | 95 | 95 | 88 | 90 | 89 |
| SVM | 76 | 94 | 93 | 82 | 79 | 80 |
| Extra Tree | 78 | 78 | 78 | 89 | 91 | 91 |
| Suggest Method (ST-GA) | 97 | 98.8 | 98 | 98 | 99.01 | 99 |

**Conclusion and future work**

diabetes has become one of the most important concerns of people and officials due to irreversible complications and its high prevalence. In this study, the PID, and Diabetes 130-US hospitals years 1999-2008 database was used to diagnose diabetes. During the last years, data mining methods have been widely used in the field of medicine and health care for diagnosing and preventing diseases and choosing treatment methods and predicting deaths and treatment costs. For this purpose, we used an ensemble learning algorithm called stacked generalization based on genetic algorithms to classify diabetic patients based on the observed complications. The purpose of this study was to combine data mining algorithms to show that combining models can improve models. According to the methods, used, the highest accuracy was obtained using the proposed Stack Generalization algorithm.

**In future research**, considering the importance of diagnosing the disease, we intend to expand the research in the field of diagnosing diseases such as breast cancer metastasis, lung cancer, Covid-19 by data mining tools and proposed algorithm and also develop and implement the subjects 1 — Consumption of drugs and supplements (drug interaction) and 2 — Provide solutions to caregivers and 3 — Introduce a specialist related to the disease and 4 — Online medical services.

**Conflict of interest**

The authors declare that no conflict of interest exists

**Acknowledgment**

We are thankful to our colleagues who provided expertise that greatly assisted the research.

**Source of funding**

All the funding of this study was provided by the authors.**Reference**

[1] Garcia-Molina, L., Lewis-Mikhael, A. M., Riquelme-Gallego, B., Cano-Ibanez, N., Oliveras-Lopez, M. J., & Bueno-Cavanillas, A. (2020). Improving type 2 diabetes mellitus glycaemic control through lifestyle modification implementing diet intervention: a systematic review and meta-analysis. European journal of nutrition, 59(4), 1313-1328.

[2] Liang, Y. Z., Li, J. J. H., Xiao, H. B., He, Y., Zhang, L., & Yan, Y. X. (2020). Identification of stress-related microRNA biomarkers in type 2 diabetes mellitus: A systematic review and meta-analysis. Journal of diabetes, 12(9), 633-644.

[3] Zhang, Y., Pan, X. F., Chen, J., Xia, L., Cao, A., Zhang, Y., ... & He, M. (2020). Combined lifestyle factors and risk of incident type 2 diabetes and prognosis among individuals with type 2 diabetes: a systematic review and meta-analysis of prospective cohort studies. Diabetologia, 63(1), 21-33.

[4] Wong, J. J., Addala, A., Abujaradeh, H., Adams, R. N., Barley, R. C., Hanes, S. J., ... & Hood, K. K. (2020). Depression in context: Important considerations for youth with type 1 vs type 2 diabetes. Pediatric diabetes, 21(1), 135-142.

[5] Australia, H. (2020). Type 2 diabetes.

[6] Abdollahi, J., Moghaddam, B. N., & Parvar, M. E. (2019). Improving diabetes diagnosis in smart health using genetic-based Ensemble learning algorithm. Approach to IoT Infrastructure. Future Gen Distrib Systems J, 1, 23-30.

[7] Wang, S., Ma, P., Zhang, S., Song, S., Wang, Z., Ma, Y., ... & Luo, H. (2020). Fasting blood glucose at admission is an independent predictor for 28-day mortality in patients with COVID-19 without previous diagnosis of diabetes: a multi-centre retrospective study. Diabetologia, 1-10.

[8] Debata, P. P., & Mohapatra, P. (2020). Diagnosis of diabetes in pregnant woman using a Chaotic-Jaya hybridized extreme learning machine model. Journal of Integrative Bioinformatics, 1(ahead-of-print).

[9] Hossain, M. S., & Muhammad, G. (2016). Cloud-assisted industrial internet of things (iiot)–enabled framework for health monitoring. Computer Networks, 101, 192-202.

[10] Temko, A. (2017). Accurate wearable heart rate monitoring during physical exercises using PPG. IEEE Transactions on Biomedical Engineering.

[11] Abawajy, J. H., & Hassan, M. M. (2017). Federated internet of things and cloud computing pervasive patient health monitoring system. IEEE Communications Magazine, 55(1), 48-53.

[12] La, H. J. (2016). A conceptual framework for trajectory-based medical analytics with IoT contexts. Journal of Computer and System Sciences, 82(4), 610-626.

[13] Rahmani, A. M., Gia, T. N., Negash, B., Anzanpour, A., Azimi, I., Jiang, M., & Liljeberg, P. (2017). Exploiting smart e-health gateways at the edge of healthcare internet-of-things: a fog computing approach. Future Generation Computer Systems.

[14] Verma, P., & Sood, S. K. (2018). Cloud-centric IoT based disease diagnosis healthcare framework. Journal of Parallel and Distributed Computing, 116, 27-38.

[15] Din, I. U., Guizani, M., Rodrigues, J. J., Hassan, S., & Korotaev, V. V. (2019). Machine learning in the Internet of Things: Designed techniques for smart cities. Future Generation Computer Systems, 100, 826-843.

[16] Fatima, M., & Pasha, M. (2017). Survey of Machine Learning Algorithms for Disease Diagnostic. Journal of Intelligent Learning Systems and Applications, 9(01), 1.

[17] Alić, B., Gurbeta, L., & Badnjević, A. (2017, June). Machine learning techniques for classification of diabetes and cardiovascular diseases. In Embedded Computing (MECO), 2017 6th Mediterranean Conference on (pp. 1-4). IEEE.

[18] Arabasadi, Z., Alizadehsani, R., Roshanzamir, M., Moosaei, H., & Yarifard, A. A. (2017). Computer aided decision making for heart disease detection using hybrid neural network-Genetic algorithm. Computer Methods and Programs in Biomedicine, 141, 19-26.

[19] Das, R., Turkoglu, I., & Sengur, A. (2009). Effective diagnosis of heart disease through neural networks ensembles. Expert systems with applications, 36(4), 7675-7680.

[20] Ekız, S., & Erdoğmuş, P. (2017, April). Comparative study of heart disease classification. In Electric Electronics, Computer Science, Biomedical Engineerings' Meeting (EBBT), 2017 (pp. 1-4). IEEE.

[21] YU-BOYUAN1,WEN-QIANGQIU1,YING-JIEWANG2,JUGAO2,PINGHE3.( 2017). CLASSIFICATION OF HEART FAILURE WITH POLYNOMIAL SMOOTH SUPPORT VECTOR MACHINE. 2017 IEEE.

[22] Beunza, J. J., Puertas, E., García-Ovejero, E., Villalba, G., Condes, E., Koleva, G., ... & Landecho, M. F. (2019). Comparison of machine learning algorithms for clinical event prediction (risk of coronary heart disease). Journal of biomedical informatics, 97, 103257.

[23] Priyanka, & Kumar, D. (2020). Decision tree classifier: a detailed survey. International Journal of Information and Decision Sciences, 12(3), 246-269.

[24] Quinlan, J. R. (1986). Induction of decision trees. Machine learning, 1(1), 81-106.

[25] Jadhav, S. D., & Channe, H. P. (2016). Comparative study of K-NN, naive Bayes and decision tree classification
۱۰